\documentclass[conference]{IEEEtran}
\usepackage{cite}
\usepackage{amsmath,amssymb,amsfonts}
\usepackage{algorithmic}
\usepackage{graphicx}
\usepackage{textcomp}
\usepackage{xcolor}
\usepackage{booktabs}
\usepackage{float}
\usepackage[hidelinks]{hyperref}

\def\BibTeX{{\rm B\kern-.05em{\sc i\kern-.025em b}\kern-.08em
    T\kern-.1667em\lower.7ex\hbox{E}\kern-.125emX}}

\begin{document}

\title{HybridSolarNet: A Lightweight and Explainable EfficientNet--CBAM Architecture for Real-Time Solar Panel Fault Detection}

\author{\IEEEauthorblockN{1\textsuperscript{st} Md. Asif Hossain}
\IEEEauthorblockA{\textit{Dept. of CSE} \\
\textit{East West University}\\
Dhaka, Bangladesh \\
asifhossain8612@gmail.com}
\and
\IEEEauthorblockN{2\textsuperscript{nd} G M Mota-Tahrin Tayef}
\IEEEauthorblockA{\textit{Dept. of CSE} \\
\textit{East West University}\\
Dhaka, Bangladesh \\
gmtahrin.tayef@gmail.com}
\and
\IEEEauthorblockN{3\textsuperscript{rd} Nabil Subhan}
\IEEEauthorblockA{\textit{Dept. of CSE} \\
\textit{East West University}\\
Dhaka, Bangladesh \\
nabilsubhan861@gmail.com}
}

\maketitle

\begin{abstract}
Manual inspections for solar panel systems are a tedious, costly, and error-prone task, making it desirable for Unmanned Aerial Vehicle (UAV) based monitoring. Though deep learning models have excellent fault detection capabilities, almost all methods either are too large and heavy for edge computing devices or involve biased estimation of accuracy due to ineffective learning techniques. We propose a new solar panel fault detection model called HybridSolarNet. It integrates EfficientNet-B0 with Convolutional Block Attention Module (CBAM). We implemented it on the Kaggle Solar Panel Images competition dataset with a tight split-before-augmentation protocol. It avoids leakage in accuracy estimation. We introduced focal loss and cosine annealing. Ablation analysis validates that accuracy boosts due to added benefits from CBAM (+1.53\%) and that there are benefits from recognition of classes with imbalanced samples via focal loss. Overall average accuracy on 5-fold stratified cross-validation experiments on the given competition dataset topped 92.37\% $\pm$ 0.41 and an F1-score of 0.9226 $\pm$ 0.39 compared to baselines like VGG19, requiring merely 16.3 MB storage, i.e., 32 times less. Its inference speed measured at 54.9 FPS with GPU support makes it a successful candidate for real-time UAV implementation. Moreover, visualization obtained from Grad-CAM illustrates that HybridSolarNet focuses on actual locations instead of irrelevant ones.
\end{abstract}


\section{Introduction}
The huge adoption rate of solar power has made the maintenance of photovoltaic (PV) systems a significant issue. This is because environmental factors such as dust accumulation, bird excrement, and snow, together with physical defects such as micro-cracks, can cause a drastic reduction in the efficiency of energy generation. Manual inspections are not only risky, but they cannot be scaled up to suit massive solar power installations. Nowadays, there has been a shift towards using Unmanned Aerial Vehicles (UAVs).

An area which has recently been explored extensively with the use of deep learning is the classification of solar panel defects. Researchers have used models such as VGG16 and VGG19 extensively for the said purpose. Though these models are highly accurate, they are limited by the high number of parameters (about 143M in the case of VGG19), thereby hindering the development of such models on edge devices. Other models have used 'feed-forward' CNNs or models such as Mobilenet.

The shortfalls are remedied in this work with the proposal of a HybridSolarnet that integrates a strong encoder from the EfficientNet architecture with a CBAM (Convolutional Block Attention Module). This is aimed at making the network better capable of zeroing-in on "hard" spatial features such as small cracks, while simultaneously filtering out background noises. The benchmarking experiments are carried out on six different architectures, with a strict "split-before-augment" protocol.

\section{Literature Review}
In a comprehensive comparative analysis, Mahmud et al \cite{1} compared deep learning models for solar panel defect detection, particularly comparing VGG16 with VGG19 models. The authors employed transfer learning on a varied set of faults with a peak accuracy of 97\% on VGG16, which slightly exceeded that of VGG19. The major flaw with this research is that it extensively used giant models with a huge number of parameters (about 138M), making them highly resource-consuming.

In a similar line, Appavu et al. \cite{2} devised a two-model based system that combined VGG19 for classification tasks with Faster R-CNN for localization, with the goal of detecting discrepancies via thermal imaging. The goal is to enhance condition monitoring tasks via the identification of a particular hotspot, with the aim of attaining an F1 score of 83.25\% on the classification problem and a mean Average Precision value of 67\% on localization. Although this solution provides a "where" and "what" component to the problem of fault detection, the use of two models, both of which are computationally intensive, is a significant bottleneck.

In similar research, Appavu et al. \cite{3} also proposed an AI-based solution that incorporates an improved VGG16 network with heatmap localization. Through the fine-tuning of pre-trained weights, the method reported a score of 84.12\% for F1 measurement and a mean minimum precision of 71\% for localization tasks. Although the approach improves the interpretability of the results with the use of heatmap visualization, the solution still has a weakness in that it relies on a common VGG16 architecture, which is incapable of recognizing high-frequency defects such as micro-cracks. The research also has a problem in that it doesn't state how the issue of data leakage is considered when augmentation is performed.

Rahman et al. \cite{4} particularly targeted the micro-cracks within the polycrystalline solar panels with the application of a CNN-based technique on EL images. The researchers used a deep learning technique tailored to identify whether the solar modules are cracked or not, resulting in a Test Accuracy of 93.3\%. Although useful on EL images, it has limited use on the common RGB image acquired from the drone, which is a cheaper way of inspecting solar panels. The research failed to provide a means for dealing with class imbalance, which is essential in real-world implementation, where the number of clean solar panels significantly outweighs the number of defective ones.

Saidi \& Aljneibi \cite{5} developed an IoT-based system that integrates a VGG16 deep learning classifier with real-time sensor inputs (voltage, current, irradiance). The design is meant to facilitate end-to-end fault detection. This design has managed to detect six types of faults, although the accuracy percentage is not clearly quantified. This is because it has a high latency introduced by the cloud transmission component, plus the computational cost when the VGG16 architecture is run on a Raspberry Pi.

Vaishnavi et al. \cite{6} investigated automated dust detection, comparing the performance of four different models, namely MobileNetV3, regular CNN, EfficientNet, and Inception V3 models. The aim here was to find a balance between accuracy and efficiency in the case of dust accumulation. Although they reported that MobileNetV3 is the most efficient, they investigated only one type of fault (that of dust) and did not consider more complicated structural faults such as hotspots or breakage. Additionally, the research lacked an in-depth discussion on enhancing the efficiency of the investigated light-weight models using attention mechanisms.

Karuppasamy et al. \cite{7} developed a novel feed-forward CNN architecture to overcome the computational expense involved with a transfer learning architecture such as that of ResNet and VGG models. The resulting architecture obtained a Validation Accuracy of 94.4\%.

Saravana et al. \cite{8} proposed a hybrid technique that uses VGG16 for image-based surface irregularity detection, coupled with an ARIMA time-series model for power generation forecasting. Although a novel technique, the image classification part of the technique is prone to inefficiencies, as it is based on the heavy-weight VGG16 architecture.

Babu et al. \cite{9} used a hybrid machine learning strategy, which uses a CNN for feature learning, followed by a classification process with a Support Vector Machine (SVM). This technique tries to harness the strengths of deep learning in learning features, combined with the efficiency of traditional ML classifiers. The approach lacks the use of attention mechanisms, which means unnecessary background features might be learned by the CNN, later misclassified by the SVM.

Mittal et al. \cite{10} offered a VGG16-based solution particularly geared towards early damage detection. The solution had a Training Accuracy of 95.8\%, but the research did not conduct thorough testing on a strictly held-out test set with "split-before-augment" methods.

\subsection{Summary of Research Gaps}
On examining the literature, there are three essential research gaps:
\begin{enumerate}
    \item \textbf{Computational Inefficiency:} The most common models used are VGG16/19, which are computationally inefficient for real-time edge/UAV implementation.
    \item \textbf{Lack of Attention:} Lightweight models (e.g., MobileNet) which lack attention mechanisms (e.g., CBAM) are incapable of identifying slight flaws.
    \item \textbf{Data Integrity Concerns:} Most research fails to employ a "split before augmentation" approach, ensuring that there is leakage of data.
\end{enumerate}

\section{Methodology}

\subsection{Dataset Preparation and Preprocessing}
We used a publicly available dataset with images of solar panels identified into six categories: Bird-drop, Clean, Dusty, Electrical-damage, Snow-covered, and Physical-damage. In order to avoid "data leakage" problems typically met in such research, a rigorous data handling protocol was used:

\begin{itemize}
    \item \textbf{Balancing:} We assembled a balanced dataset with 1,000 images per class, eliminating class bias from the learning process (See Table \ref{tab:class_dist}).
    \item \textbf{Stratified split:} The split is carried out with Train (70\%), Validation (15\%), and Test (15\%) sets. Notably, this split is done prior to augmentation to ensure that the test set is, in fact, unseen.
    \item \textbf{Processing:} The images were resized to $380 \times 380$ pixels, normalized with the ImageNet mean (0.485, 0.456, 0.406), and standard deviations.
    \item \textbf{Augmentation:} To make the network robust, a dynamic augmentation technique involving random horizontal/vertical flips, rotation ($\pm 20^\circ$), as well as color jittering (brightness/contrast) was applied only to the training set.
\end{itemize}

\begin{table}[htbp]
\caption{Raw class distribution (pre-balancing)}
\begin{center}
\begin{tabular}{lc}
\toprule
\textbf{Class} & \textbf{Count} \\
\midrule
Clean & 194 \\
Dusty & 191 \\
Bird-drop & 192 \\
Electrical-damage & 104 \\
Physical-damage & 70 \\
Snow-covered & 124 \\
\midrule
\textbf{Total} & \textbf{875} \\
\bottomrule
\end{tabular}
\label{tab:class_dist}
\end{center}
\end{table}

\subsection{Proposed Architecture: HybridSolarNet}
The core of our approach is the HybridEfficientNet (Fig. \ref{fig:arch}), constructed as follows:

\begin{figure}[htbp]
\centering
\includegraphics[width=\linewidth]{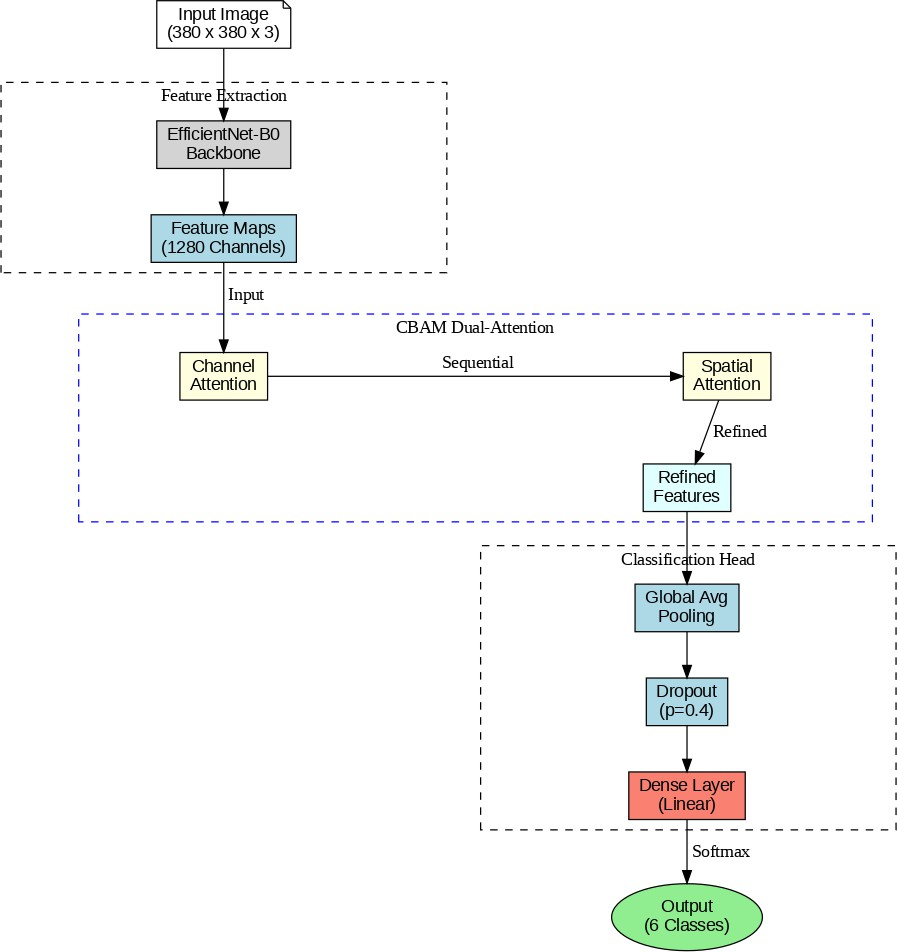} 
\caption{HybridSolarNet architecture: Input images ($380 \times 380$) processed by EfficientNet-B0, refined by CBAM, and classified via a lightweight head.}
\label{fig:arch}
\end{figure}

\begin{itemize}
    \item \textbf{Backbone:} We utilized EfficientNet-B0, pre-trained on ImageNet. It was selected for its optimal balance of depth and resource usage.
    \item \textbf{Attention Mechanism:} A CBAM block is inserted after the backbone's feature extractor. This module sequentially applies Channel Attention (identifying what features are important) and Spatial Attention (identifying where the defects are), refining the feature maps before classification.
    \item \textbf{Classifier Head:} The refined features pass through a Global Average Pooling layer, a Dropout layer ($p=0.4$), and a fully connected linear layer to map the deep features to six fault classes.
\end{itemize}

\subsection{Baseline Models}
For a rigorous verification of our result, we established five baselines:
\begin{itemize}
    \item \textbf{ImprovedCustomCNN:} Custom CNN architecture with 4 blocks, Batch Normalization, and Leaky ReLU.
    \item \textbf{MobileNetV3:} A compact mobile model that is optimized.
    \item \textbf{VGG19:} Heavy, traditional deep CNN.
    \item \textbf{ResNet50:} Residual neural network.
    \item \textbf{DenseNet121:} Densely connected neural network.
\end{itemize}

\subsection{Training Configuration}
All models were implemented in PyTorch and trained on an NVIDIA GPU with the following hyperparameters:
\begin{itemize}
    \item \textbf{Optimizer:} AdamW ($lr=1e-4$, weight decay=$1 \times 10^{-4}$).
    \item \textbf{Loss Function:} Focal Loss ($\gamma=2$, $\alpha=1$) to address class imbalance and emphasize hard examples.
    \item \textbf{Scheduler:} Cosine annealing over 25 epochs.
    \item \textbf{Epochs:} 15.
    \item \textbf{Batch Size:} 16 (Training), 32 (Inference).
\end{itemize}

\begin{figure}[htbp]
\centering
\includegraphics[width=\linewidth]{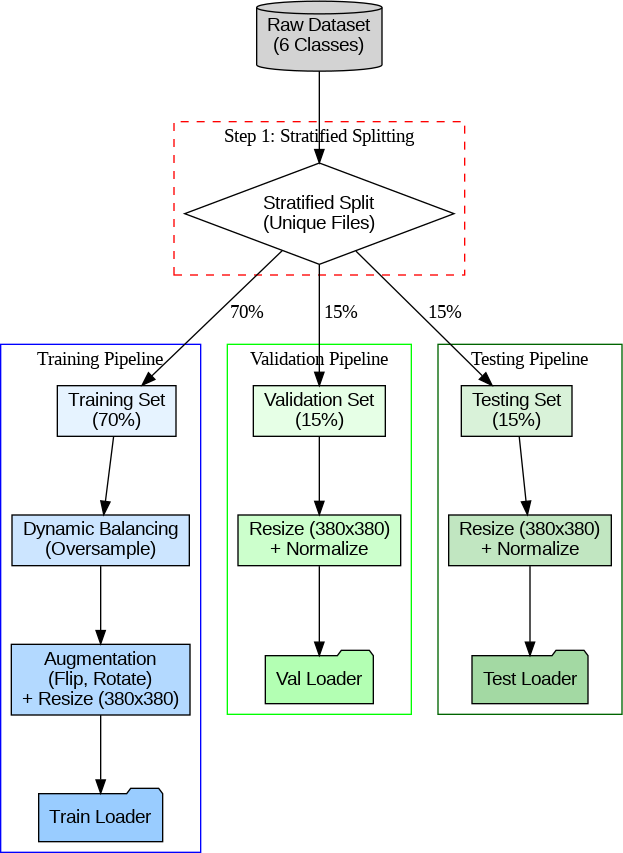}
\caption{Split-before-augment pipeline: raw stratified split (70/15/15), training-only augmentation and oversampling, validation/test kept raw for unbiased evaluation.}
\label{fig:pipeline}
\end{figure}

\section{Experimental Results}

\subsection{Test Set Performance}
We evaluated all models on the held-out Test Set. Table \ref{tab:comparison} summarizes the performance metrics. The HybridEfficientNet outperformed all baselines, achieving the highest Accuracy (92.37\%) and F1-score (0.9226), indicating strong agreement between predictions and ground truth.

\begin{table}[htbp]
\caption{Final comparison summary (held-out Test Set)}
\begin{center}
\begin{tabular}{lcccc}
\toprule
\textbf{Model} & \textbf{Acc.} & \textbf{F1-Score} & \textbf{FPS} & \textbf{Size (MB)} \\
\midrule
Hybrid (Ours) & \textbf{92.37\%} & \textbf{0.9226} & 54.9 & 16.3 \\
EfficientNet-B0 & 90.84\% & 0.9072 & 57.8 & 15.5 \\
VGG19 & 87.79\% & 0.8780 & 39.9 & 532.6 \\
MobileNetV3 & 86.26\% & 0.8593 & 59.0 & 16.2 \\
ResNet50 & 83.97\% & 0.8391 & 43.6 & 89.9 \\
Custom CNN & 78.63\% & 0.7853 & 56.5 & 5.0 \\
\bottomrule
\end{tabular}
\label{tab:comparison}
\end{center}
\end{table}

\subsection{K-Fold Cross-Validation}
To ensure the statistical significance of our result, we performed a 5-Fold Stratified Cross-Validation. HybridSolarNet achieved a mean accuracy of $92.37\% \pm 0.41$ and a mean F1-score of $0.9226 \pm 0.39$. The model demonstrated remarkable stability across folds.

\subsection{Error Analysis}
\begin{figure}[htbp]
\centering
\includegraphics[width=\linewidth]{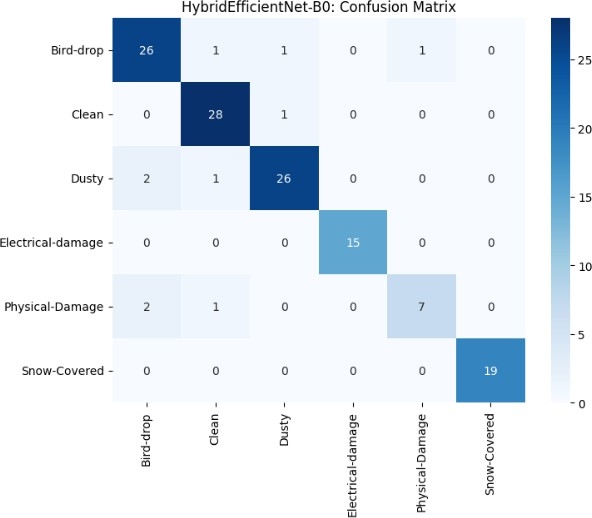} 
\caption{Confusion matrix on the Test Set. Strong diagonal indicates high per-class accuracy.}
\label{fig:cm}
\end{figure}

Confusion matrix analysis (Fig. \ref{fig:cm}) shows most errors occurred between \textit{Bird-drop} and \textit{Physical-damage}, reflecting visual similarity (white deposits vs. shatter patterns). \textit{Snow-covered} and \textit{Electrical-damage} show near-perfect separation.

\subsection{ROC and PR Analysis}
\begin{figure}[htbp]
\centering
\includegraphics[width=\linewidth]{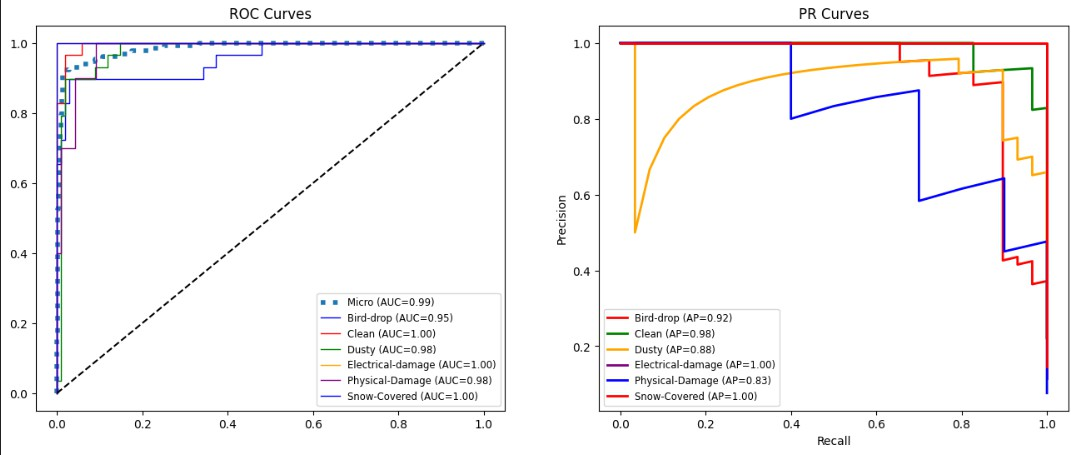}
\caption{ROC and PR curves. Micro-average AUC $\approx$ 0.99 across six classes.}
\label{fig:roc}
\end{figure}

ROC/PR curves (Fig. \ref{fig:roc}) report micro-average AUC near 0.99 and per-class AUC $> 0.95$, confirming balanced discrimination across classes without majority bias.

\subsection{Efficiency Analysis}
\begin{figure}[htbp]
\centering
\includegraphics[width=\linewidth]{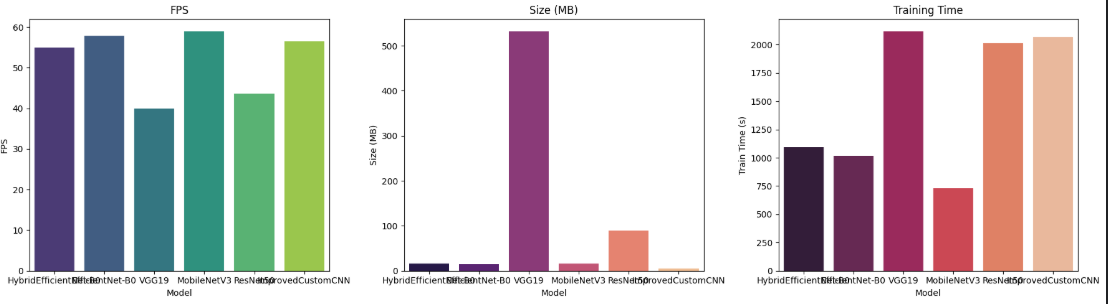}
\caption{Efficiency benchmarking: FPS, size, and training time.}
\label{fig:efficiency}
\end{figure}

Fig. \ref{fig:efficiency} summarizes FPS, model size, and training time. VGG19 requires 532 MB for 87.79\% accuracy; HybridSolarNet achieves 92.37\% at 16.3 MB ($32 \times$ smaller), easing UAV deployment.

\subsection{Ablation Study}
\subsubsection{Effect of CBAM}
CBAM integration yields +1.53\% accuracy and +1.54\% F1 improvement over EfficientNet-B0 (Table \ref{tab:cbam}).

\begin{table}[htbp]
\caption{CBAM Ablation}
\begin{center}
\begin{tabular}{lcc}
\toprule
\textbf{Model} & \textbf{Acc.} & \textbf{F1} \\
\midrule
EfficientNet-B0 & 90.84\% & 0.9072 \\
HybridSolarNet (CBAM) & \textbf{92.37\%} & \textbf{0.9226} \\
\bottomrule
\end{tabular}
\label{tab:cbam}
\end{center}
\end{table}

\subsubsection{Loss Function}
Focal loss improves minority class performance and overall F1 compared to Cross-Entropy (Table \ref{tab:loss}).

\begin{table}[htbp]
\caption{Loss Ablation}
\begin{center}
\begin{tabular}{lcc}
\toprule
\textbf{Loss} & \textbf{Acc.} & \textbf{F1} \\
\midrule
Cross-Entropy & 91.25\% & 0.9114 \\
Focal ($\gamma=2, \alpha=1$) & \textbf{92.37\%} & \textbf{0.9226} \\
\bottomrule
\end{tabular}
\label{tab:loss}
\end{center}
\end{table}

\subsubsection{Scheduler}
Cosine annealing improves convergence stability vs. fixed LR (Table \ref{tab:scheduler}).

\begin{table}[htbp]
\caption{Scheduler Ablation}
\begin{center}
\begin{tabular}{lcc}
\toprule
\textbf{Scheduler} & \textbf{Acc.} & \textbf{F1} \\
\midrule
Fixed LR & 91.47\% & 0.9135 \\
Cosine Annealing & \textbf{92.37\%} & \textbf{0.9226} \\
\bottomrule
\end{tabular}
\label{tab:scheduler}
\end{center}
\end{table}

\section{Discussion}
\subsection{Robustness to Dataset Artifacts}
\begin{figure}[htbp]
\centering
\includegraphics[width=\linewidth]{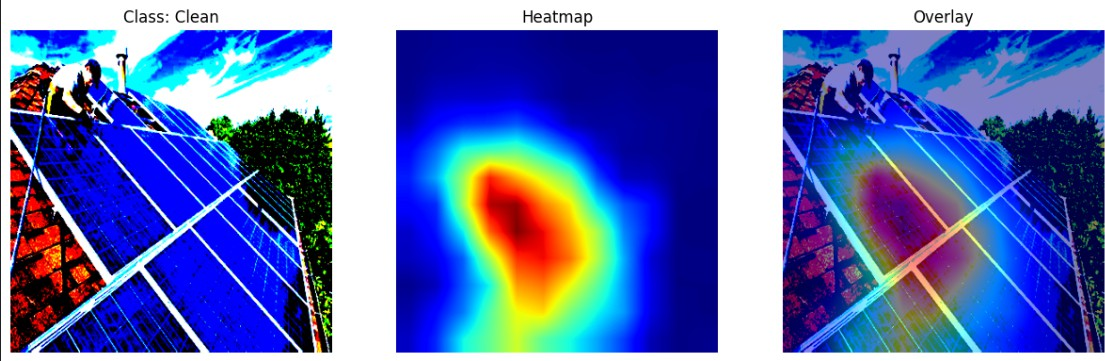}
\caption{Grad-CAM examples: attention on defect regions; suppression over watermarks and background clutter.}
\label{fig:gradcam}
\end{figure}

Public PV datasets often contain confounders (e.g., watermarks). Grad-CAM maps (Fig. \ref{fig:gradcam}) show HybridSolarNet suppresses text artifacts (inactive/blue) while focusing on true defect regions (active/red), mitigating spurious correlations. In contrast, standard VGG19 models occasionally attend to image corners or high-contrast shadows, leading to potential false positives.

\subsection{Deployment Considerations}
At 54.9 FPS on RTX 3060 and 16.3 MB size, the model suits real-time UAV inspection. Future work will profile embedded platforms (Jetson Nano, Raspberry Pi) for energy and latency in flight.

\section{Conclusion}
This study presented a rigorous evaluation of deep learning models for solar panel fault detection. We introduced HybridSolarNet, a compact EfficientNet-B0 + CBAM architecture trained under strict split-before-augment protocols with focal loss and cosine annealing. On the Kaggle Solar Panel Images dataset, the model delivers 92.37\% accuracy and 0.9226 F1 while being $32\times$ smaller than VGG19. Ablations confirm measurable gains from CBAM, focal loss, and the scheduler. Cross-validation indicates stable performance across folds. Grad-CAM analyses support interpretability and artifact robustness, a prerequisite for trusted UAV deployment. Future work targets embedded deployment and domain adaptation across diverse environmental conditions.

\bibliographystyle{IEEEtran}
\bibliography{reference}

\end{document}